\documentclass[10pt,journal,compsoc]{IEEEtran}

\usepackage{url}
\usepackage[utf8]{inputenc}
\usepackage{acronym}

\usepackage{subcaption}
\usepackage{amsmath}
\usepackage{graphicx} % more modern
\usepackage{color}
\usepackage{ifthen}
\usepackage{balance}
\usepackage{amssymb}

\acrodef{CNN}{Convolutional Neural Network}
\acrodef{DRAW}{Deep Recurrent Attentive Writer}
\acrodef{GRU}{Gated Recurrent Unit}
\acrodef{IoU}{Intersection-over-Union}
\acrodef{LSTM}{Long Short-Term Memory}
\acrodef{MSE}{Mean Squared Error}
\acrodef{RATM}{Recurrent Attentional Tracking Model}
\acrodef{ReLU}{Rectified Linear Unit}
\acrodef{RNN}{Recurrent Neural Network}
\acrodef{SGD}{Stochastic Gradient Descent}
\ifCLASSOPTIONcompsoc
  \usepackage[nocompress]{cite}
\else
  \usepackage{cite}
\fi
\ifCLASSINFOpdf
\else
\fi
\begin{document}
\title{RATM: Recurrent Attentive Tracking Model}

\author{Samira Ebrahimi Kahou,~Vincent Michalski, and Roland Memisevic%
%Michael~Shell,~\IEEEmembership{Member,~IEEE,}
%        John~Doe,~\IEEEmembership{Fellow,~OSA,}
%        and~Jane~Doe,~\IEEEmembership{Life~Fellow,~IEEE}% <-this % stops a space
%\thanks{M. Shell was with the Department
%of Electrical and Computer Engineering, Georgia Institute of Technology, Atlanta,
}

\IEEEtitleabstractindextext{
\begin{abstract}
We present an attention-based modular neural framework for computer vision.
The framework uses a soft attention mechanism allowing models to be trained with gradient descent.
It consists of three modules:
a recurrent attention module controlling \emph{where} to look in an image or video frame,
a feature-extraction module providing a representation of \emph{what} is seen, and
an objective module formalizing \emph{why} the model learns its attentive behavior.
The attention module allows the model to focus computation on task-related information in the input. 
We apply the framework to several object tracking tasks and explore various design choices.
We experiment with three data sets, bouncing ball, moving digits and the real-world KTH data set.
The proposed \acf{RATM} performs well on all three tasks and can generalize to related but previously unseen sequences from a challenging tracking data set. 
\end{abstract}
\acresetall

}

\maketitle

\IEEEdisplaynontitleabstractindextext

% \fi
\IEEEpeerreviewmaketitle

\IEEEraisesectionheading{\section{Introduction}\label{sec:introduction}}
\IEEEPARstart{A}{ttention mechanisms} are one of the biggest trends in deep-learning research and
have been successfully applied in a variety of neural-network architectures across different tasks.
In computer vision, for instance, attention mechanisms have been used for image generation \cite{DRAW} and image captioning \cite{xu2015show}.
In natural language processing they have been used for machine translation \cite{bahdanau2014neural} and sentence summarization \cite{rush2015neural}. 
And in computational biology attention was used for subcellular protein localization \cite{sonderby2015convolutional}.

In these kinds of applications usually not all information contained in the input data is relevant for the given task. 
Attention mechanisms allow the neural network to focus on the relevant parts of the input, while
ignoring other, potentially distracting, information.
Besides enabling models to ignore distracting information, attention mechanisms can be helpful in streaming data scenarios, where the amount of data per frame can be prohibitively large for full processing.
In addition, some studies suggest that there is a representational advantage of sequential processing of image parts over a single pass over the whole image (see for example~\cite{mnih2014recurrent,larochelle2010learning,DRAW,denil2011,ranzato2014learning,sermanet2014attention}).

Recently, \cite{DRAW} introduced the \ac{DRAW}, which involves a \acf{RNN} that controls a read and a write mechanism based on attention. The read mechanism extracts a parametrized window from the static input image.
Similarly, the write mechanism is used to write into a window on an output canvas.
This model is trained to sequentially produce a reconstruction of the input image on the canvas.
Interestingly, one of the experiments on handwritten digits showed that the read mechanism learns to trace digit contours and the write mechanism generates digits in a continuous motion.
This observation hints at the potential of such mechanisms in visual object tracking applications, where the primary goal is to trace the spatio-temporal ``contours'' of an object as it moves in a video.

Previous work on the application of attention mechanisms for tracking includes \cite{denil2011} and references therein. 
In contrast to that line of work, we propose a model based on a fully-integrated neural framework, that can be trained end-to-end using back-propagation.
The framework consists of three modules: a recurrent differentiable attention module controlling \emph{where} to look in an image,
a feature-extraction module providing a representation of \emph{what} is seen, and
an objective module formalizing \emph{why} the model learns its attentive behavior.
As we shall show, a suitable surrogate cost in the objective module can provide a supervised learning signal, that allows us to train the network end-to-end, and to learn attentional strategies using simple supervised back-prop without resorting to reinforcement learning or sampling methods.

According to a recent survey of tracking methods \cite{smeulders2014visual}, many approaches to visual tracking involve a search over multiple window candidates based on a similarity measure in a feature space.
Successful methods involving deep learning, such as \cite{nam2015mdnet}, perform tracking-by-detection, e.g. by using a \ac{CNN} for foreground-background classification of region proposals. 
As in most approaches, the method in \cite{nam2015mdnet} at each time step samples a number of region proposals ($256$) from a Gaussian distribution centered on the region of the previous frame.
Such methods do not benefit from useful correlations between the target location and the object's past trajectory.
There are deep-learning approaches that consider trajectories by employing particle filters such as  \cite{wang2013learning}, which still involves ranking of region proposals ($1,000$ particles).

In our \ac{RATM}, an \ac{RNN} predicts the position of an object at time $t$, given a real-valued hidden state vector. The state vector can summarize the history of observations and predictions of previous time steps.
We rely on a \emph{single} prediction per time step instead of using the predicted location as basis for a search over multiple region proposals.
This allows for easy integration of our framework's components and training with simple gradient-based methods.

The main contribution of our work is the introduction of a modular neural framework, that can be trained end-to-end with gradient-based learning methods.
Using object tracking as an example application, we explore different settings and provide insights into model design and training.
While the proposed framework is targeted primarily at videos, it can also be applied to sequential processing of still images.

\section{Recurrent Neural Networks}
\label{sec:rnn}
\begin{figure}[h]
    \centering
    \begin{subfigure}{.3\columnwidth}
        \centering
        \includegraphics[width=.99\textwidth]{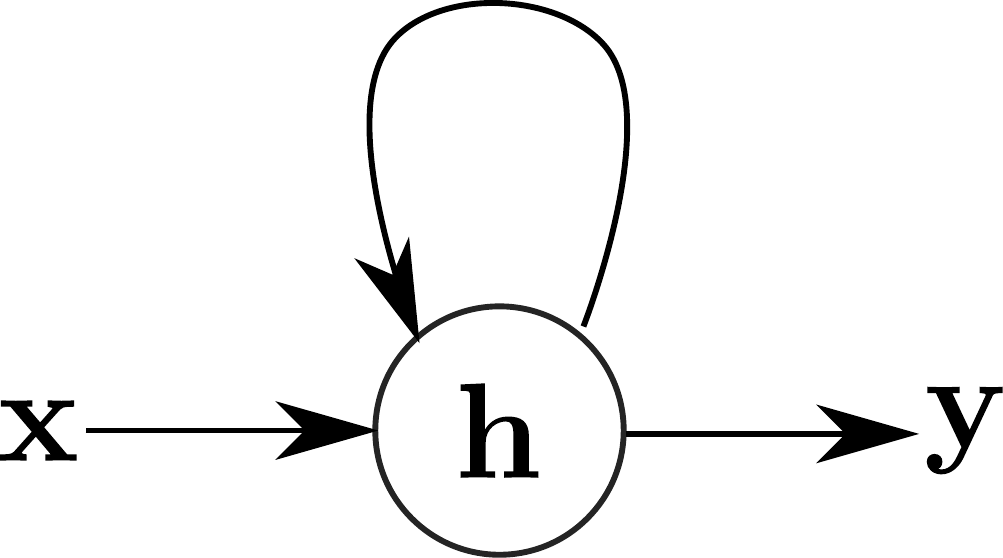}
    \end{subfigure}
    \hspace{.1\columnwidth}
    \begin{subfigure}{.4\columnwidth}
        \centering
        \includegraphics[width=.99\textwidth]{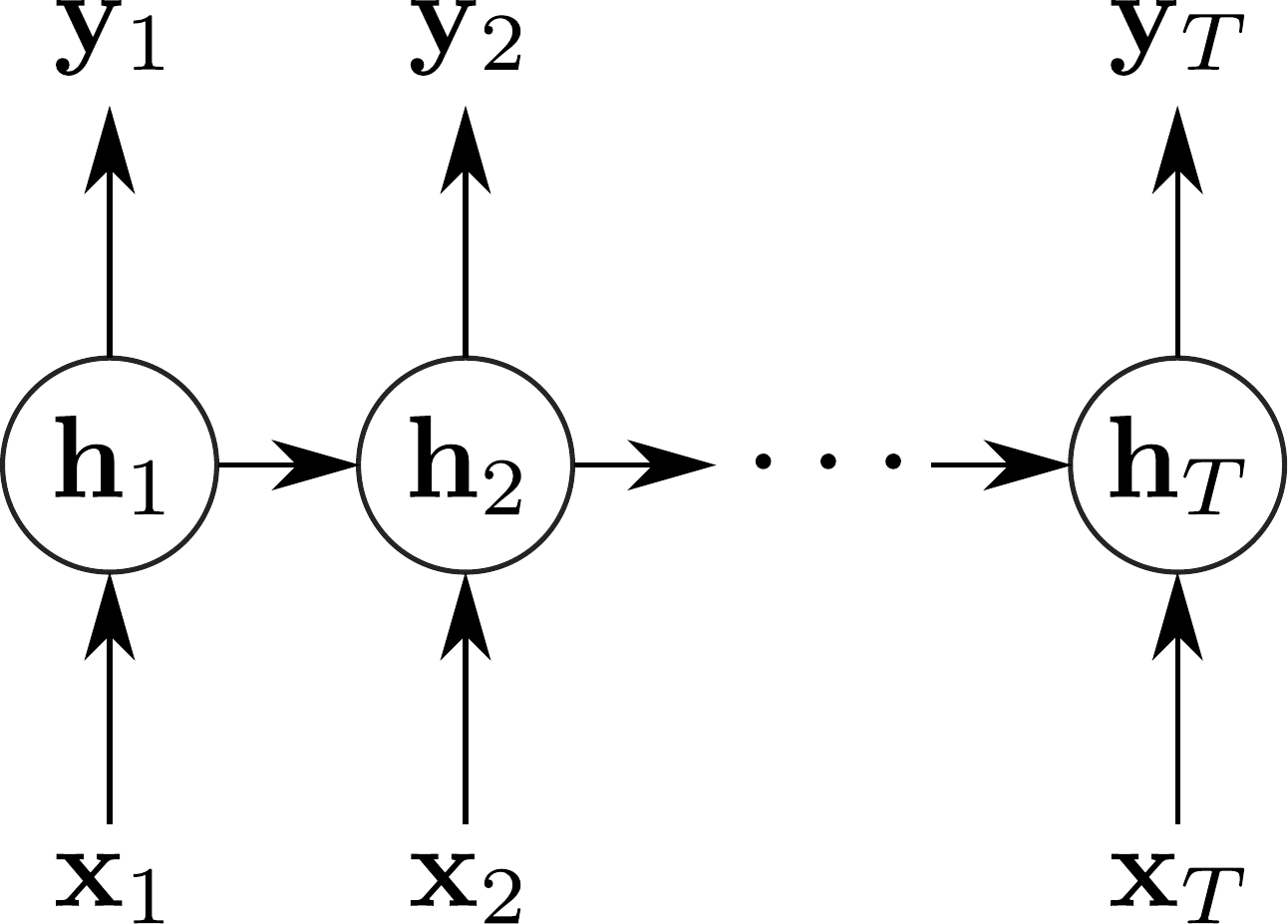}
    \end{subfigure}
    \caption{Left: A simple recurrent network with one input, one output and a hidden layer, which has a feed-back loop connection.
        Right: The same network unrolled in time for $T$ time steps.}
    \label{fig:rnn}
\end{figure}

\acfp{RNN} are powerful machine learning models that are used for learning in sequential processing tasks.
Advances in understanding the learning dynamics of \acp{RNN} enabled their successful application in a wide range of tasks (for example~\cite{hochreiter1997long,pascanu2012difficulty,graves2013speech,sutskever2014sequence,cho2014learning,srivastava2015unsupervised}). 
The standard \ac{RNN} model consists of an input, a hidden and an output layer as illustrated in Figure \ref{fig:rnn}.

In each time step $t$, the network computes a new hidden state $\mathbf{h}_{t}$ based on the previous state $\mathbf{h}_{t-1}$ and the input $\mathbf{x}_t$:
\begin{equation}
\label{eq:rnn_hid}
\mathbf{h_t} = \sigma(\mathbf{W}_{in}\mathbf{x}_t + \mathbf{W}_{rec}\mathbf{h}_{t-1}),
\end{equation}
where $\sigma$ is a non-linear activation function, $\mathbf{W}_{in}$ is the matrix containing the input-to-hidden weights and $\mathbf{W}_{rec}$ is the recurrent weight matrix from the hidden layer to itself.
At each time step the \ac{RNN} also generates an output
\begin{equation}
    \label{eq:rnn_out}
    \mathbf{y}_t = \mathbf{W}_{out}\mathbf{h}_t + \mathbf{b}_y,
\end{equation}
where $\mathbf{W}_{out}$ is the matrix with weights from the hidden to the output layer.

Although the application of recurrent networks with sophisticated hidden units, such as \ac{LSTM}~\cite{hochreiter1997long} or \ac{GRU}~\cite{cho2014learning}, has become common in recent years (for example~\cite{bahdanau2014neural,sutskever2014sequence,srivastava2015unsupervised}),
we rely on the simple IRNN proposed by~\cite{le2015simple}, and show that it works well in the context of visual attention. 
The IRNN corresponds to a standard \ac{RNN}, where recurrent weights $\mathbf{W}_{rec}$ are initialized with a scaled version of the identity matrix and the hidden activation function $\sigma(.)$ is the element-wise \ac{ReLU} function~\cite{nair2010rectified}
\begin{equation}
\sigma(x) = max(0, x). 
\end{equation}
The initial hidden state $\mathbf{h}_0$ is initialized as the zero vector.
Our experiments are based on the Theano~\cite{bastien2012theano,bergstra2010theano} implementation of the IRNN shown to work well for video in~\cite{kahou2015recurrent}.

\section{Neural Attention Mechanisms}
\label{sec:attention}
\begin{figure}[h]
    \begin{center}
        \includegraphics[width=.8\columnwidth]{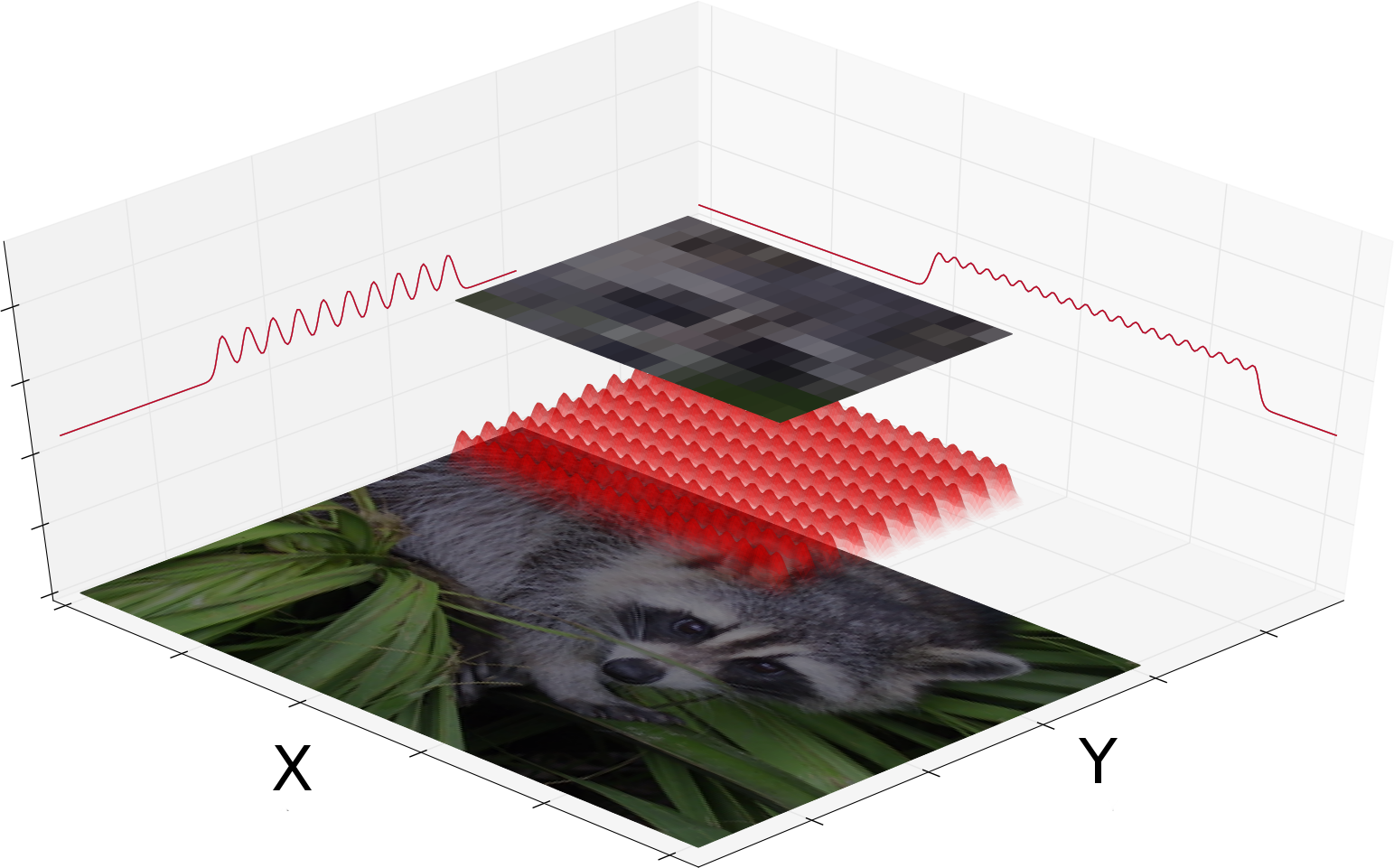}
    \end{center}
    \caption{A $20\times10$ glimpse is extracted from the full image by applying a grid of $20\times10$ two-dimensional Gaussian window filters. 
    The separability of the multi-dimensional Gaussian window allows for efficient computation of the extracted glimpse.}
    \label{fig:patch_extraction}
\end{figure}

Our attention mechanism is a modification of the read mechanism introduced in~\cite{DRAW}. It extracts \emph{glimpses} from the input image by applying a grid of two-dimensional Gaussian window filters. Each of the filter responses corresponds to one pixel of the glimpse.
An example of the glimpse extraction is shown in Figure~\ref{fig:patch_extraction}.

Given an image $\mathbf{x}$ with $A$ columns and $B$ rows, the attention mechanism separately applies a set of $M$ column filters $\mathbf{F}_X\in\mathbb{R}^{M\times A}$ and a set of $N$ row filters $\mathbf{F}_Y\in\mathbb{R}^{N\times B}$, extracting an $M\times N$ glimpse $\mathbf{p}$:
\begin{equation}
\mathbf{p} = \mathbf{F}_Y\mathbf{x}\mathbf{F}_{X}^\mathrm{T}.
\end{equation}
This implicitly computes $M\times N$ two-dimensional filter responses due to the separability of two-dimensional Gaussian filters. For multi-channel images the same filters are applied to each channel separately.

The sets of one-dimensional row ($\mathbf{F}_Y$) and column ($\mathbf{F}_X$) filters have three parameters each\footnote{The original read mechanism in \cite{DRAW} also adds a scalar intensity parameter $\gamma$, that is multiplied to filter responses.}:
\begin{itemize}
    \item the grid center coordinates $g_X, g_Y$,
    \item the standard deviation for each axis $\sigma_X, \sigma_Y$ and
    \item the stride between grid points on each axis $\delta_X, \delta_Y$.
\end{itemize}

These parameters are dynamically computed as an affine transformation of a vector of activations $\mathbf{h}$ from a neural network layer:
\begin{equation}
\label{eq:att1}
(\tilde{g}_X, \tilde{g}_Y, \tilde{\sigma_X}, \tilde{\sigma_Y}, \tilde{\delta_X}, \tilde{\delta}_Y) = \mathbf{W}\mathbf{h} + \mathbf{b},
\end{equation}
where $\mathbf{W}$ is the transformation matrix and $\mathbf{b}$ is the bias.
This is followed by normalization of the parameters:
\begin{align}
\label{eq:norm_centers}
g_X &= \frac{\tilde{g}_X + 1}{2},&\,\,g_Y &= \frac{\tilde{g}_Y + 1}{2},\\
\label{eq:norm_strides}
\delta_X &= \frac{A - 1}{M-1} \cdot |\tilde{\delta_X}|,&\,\,
\delta_Y &= \frac{B - 1}{N-1} \cdot |\tilde{\delta_Y}|,\\
\label{eq:norm_stds}
\sigma_X &= |\tilde{\sigma_X}|,&\,\,
\sigma_Y &= |\tilde{\sigma_Y}|.
\end{align}

The mean coordinates $\mu^{i}_X, \mu^{j}_Y$ of the Gaussian filter at column $i$, row $j$ in the attention grid are computed as follows:
\begin{align}
\mu^{i}_X &= g_X + (i - \frac{M}{2} - 0.5)\cdot\delta_X,\\
\mu^{j}_Y &= g_Y + (j - \frac{N}{2} - 0.5)\cdot\delta_Y
\end{align}

Finally, the filter banks $\mathbf{F}_X$ and $\mathbf{F}_Y$ are defined by:
\begin{align}
    \label{eq:filter}
    \mathbf{F}_X[i,a] &= \exp\left(-\frac{(a-\mu_{X}^i)^2}{2\sigma^2}\right),\\
    \mathbf{F}_Y[j,b] &= \exp\left(-\frac{(b-\mu_{Y}^j)^2}{2\sigma^2}\right)
\end{align}
The filters (rows of $\mathbf{F}_X$ and $\mathbf{F}_Y$) are later normalized to sum to one.

Our read mechanism makes the following modifications to the \ac{DRAW} read mechanism~\cite{DRAW}:
\begin{itemize}
    \item We allow rectangular (not only square) attention grids and we use separate strides and standard deviations for the $X$ and $Y$-axis. This allows the model to stretch and smooth the glimpse content to correct for distortions introduced by ignoring the original aspect ratio of an input image.
    \item We use the absolute value function $abs(x) = |x|$ instead of $\exp(x)$ to ensure positivity of strides and standard deviations (see Equations~\ref{eq:norm_strides} and \ref{eq:norm_stds}).
The motivation for this modification is that in our experiments we observed stride and standard deviation parameters to often saturate at low values, causing the attention window to zoom in on a single pixel. This effectively inhibits gradient flow through neighboring pixels of the attention filters. 
Piecewise linear activation functions have been shown to benefit optimization \cite{nair2010rectified} and the absolute value function is a convenient trade-off between the harsh zeroing of all negative inputs of the \ac{ReLU} and the extreme saturation for highly negative inputs of the exponential function.
    \item We drop the additional scalar intensity parameter $\gamma$, because we did not observe it to influence the performance in our experiments.
\end{itemize}

\section{A Modular Framework for Vision}
\label{sec:modular_framework}
The proposed modular framework for an attention-based approach to computer vision consists of three components: 
the attention module (controlling \emph{where} to look), the feature-extraction module (providing a representation of \emph{what} is seen) and the objective module (formalizing \emph{why} the model is learning its attentive behavior). 
An example architecture for tracking using these modules is described in Section~\ref{sec:ratm}.

\subsection{Feature-extraction module}
The feature-extraction module computes a feature representation of a given input glimpse. 
This representation can be as simple as the identity transformation, i.e. the original pixel representation, or a more sophisticated feature extractor, e.g. an \ac{CNN}.
The extracted features are used by other modules to reason about the visual input. Given a hierarchy of features, such as the activations of layers in an \ac{CNN}, different features can be passed to the attention and objective modules.

We found that it can be useful to pre-train the feature-extraction module on a large data set, before starting to train the full architecture. After pre-training, the feature extractor's parameters can either be continued to be updated during end-to-end training, or kept fixed. 

\begin{figure}[h]
    \centering
    \includegraphics[width=0.7\columnwidth]{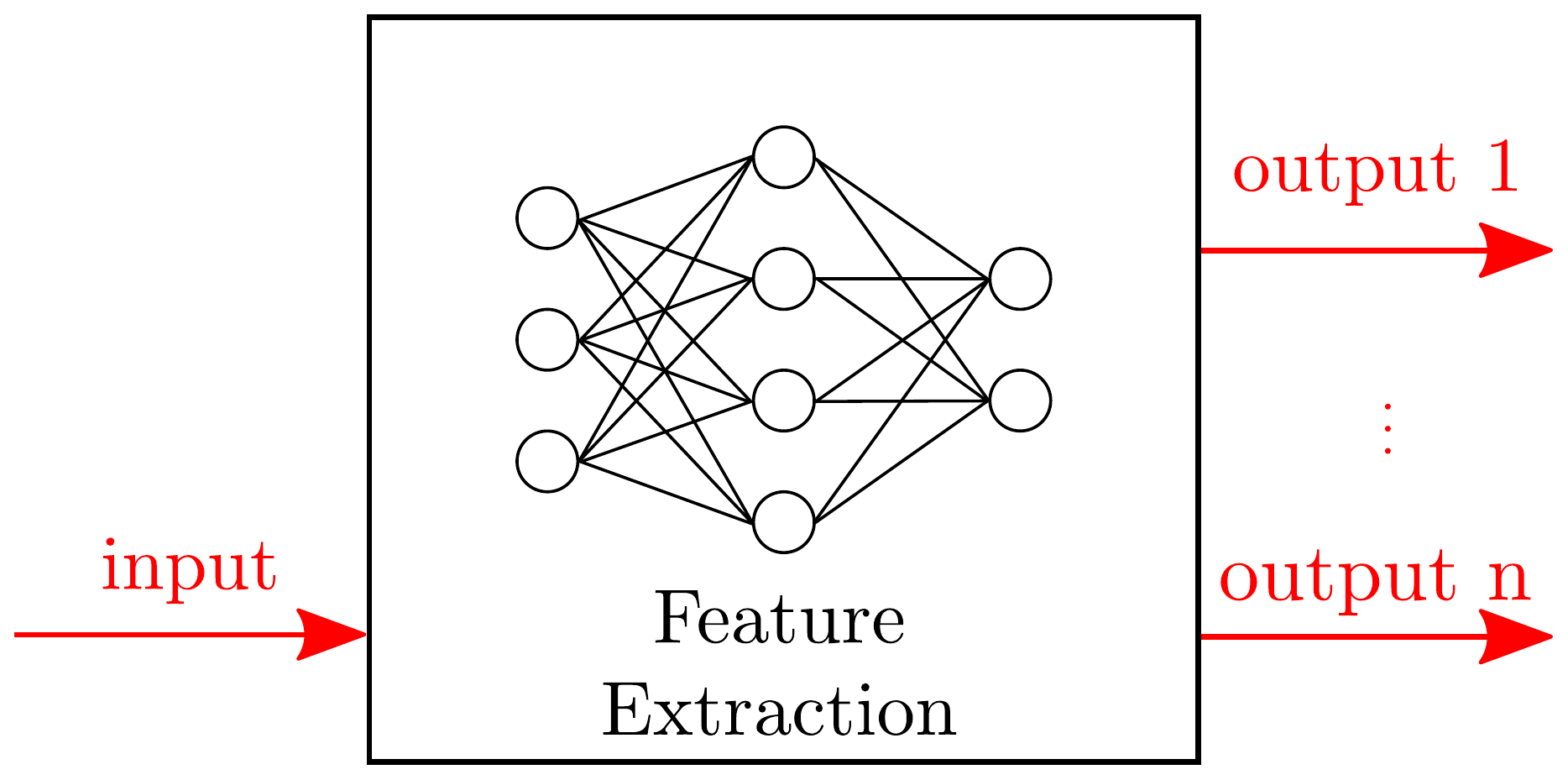}
    \caption{The symbol for the feature-extraction module. It represents an arbitrary feature extractor, that can have multiple outputs (e.g. for activations from different layers of an \ac{CNN}).}
    \label{fig:feat_extr_mod}
\end{figure}

Figure~\ref{fig:feat_extr_mod} shows the symbol used in the following sections to represent a feature-extraction module.

\subsection{Attention Module}
\begin{figure}[h]
    \centering
    \includegraphics[width=0.99\columnwidth]{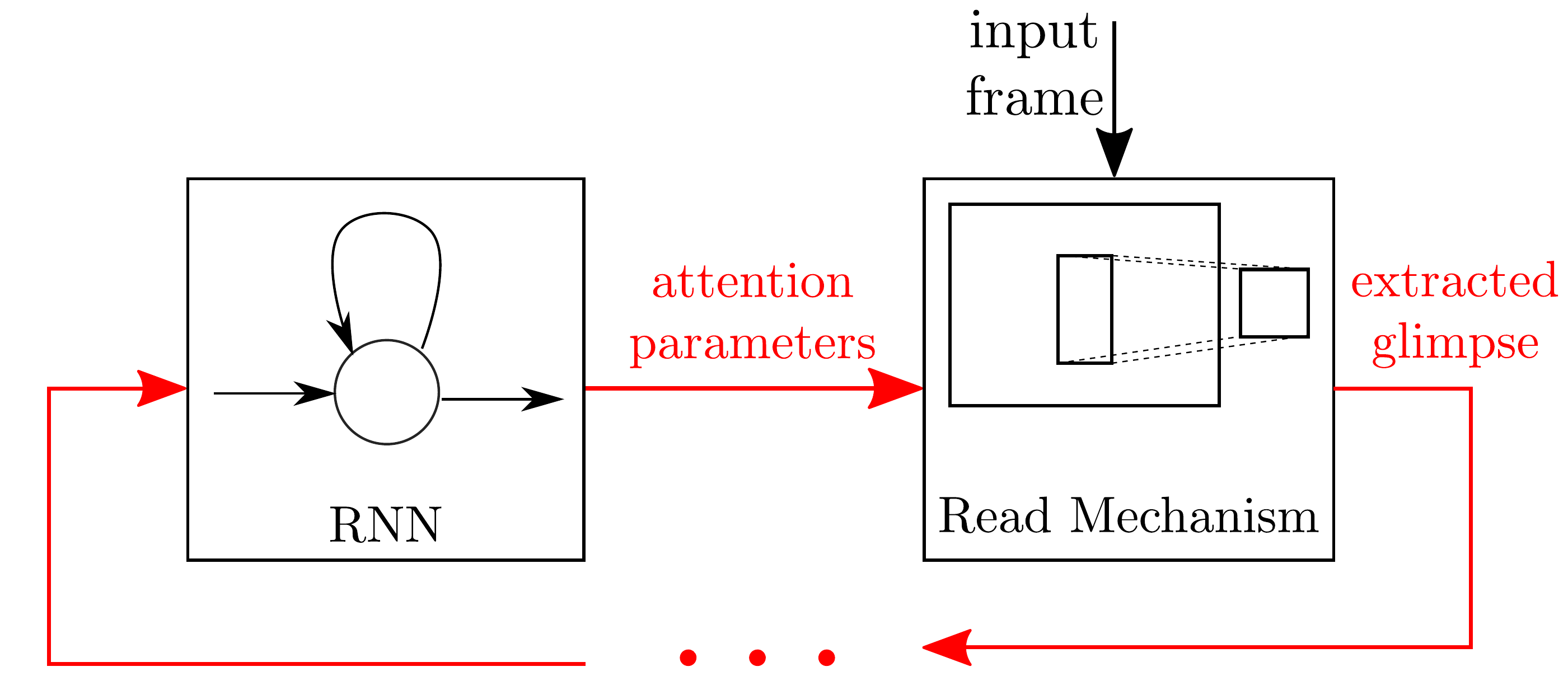}
    \caption{The symbolic representation of a recurrent attention module, which is composed of an \ac{RNN} and a read mechanism that extracts a glimpse from the input frame. The extracted glimpse is fed back to the \ac{RNN}. The dots indicate, that the feed-back connection can involve intermediate processing steps, such as feature extraction.}
    \label{fig:rec_att_mod}
\end{figure}

The attention module is composed of an \ac{RNN} (see Section~\ref{sec:rnn}) and a read mechanism (see Section~\ref{sec:attention}). 
At each time step, a glimpse is extracted from the current input frame using the attention parameters the \ac{RNN} predicted in the previous time step (see Section~\ref{sec:attention}). 
Note that in this context, Equation~\ref{eq:att1} of the read mechanism corresponds to Equation~\ref{eq:rnn_out} of the \ac{RNN}.
After the glimpse extraction, the \ac{RNN} updates its hidden state using the feature representation of the glimpse as input (see Equation~\ref{eq:rnn_hid}).
Figure~\ref{fig:rec_att_mod} shows the symbolic representation used in the following sections to represent the recurrent attention module.

\subsection{Objective Module}
An objective module guides the model to learn an attentional policy to solve a given task.
It outputs a scalar cost, that is computed as function of its target and prediction inputs.
There can be multiple objective modules for a single task. 
A learning algorithm, such as \ac{SGD}, uses the sum of cost terms from all objective modules to adapt the parameters of the other modules. 
Objective modules can receive their input from different parts of the network. 
For example, if we want to define a penalty between window coordinates, the module would receive predicted attention parameters from the attention module and target parameters from the trainer.

\begin{figure}[h]
    \centering
    \includegraphics[width=0.7\columnwidth]{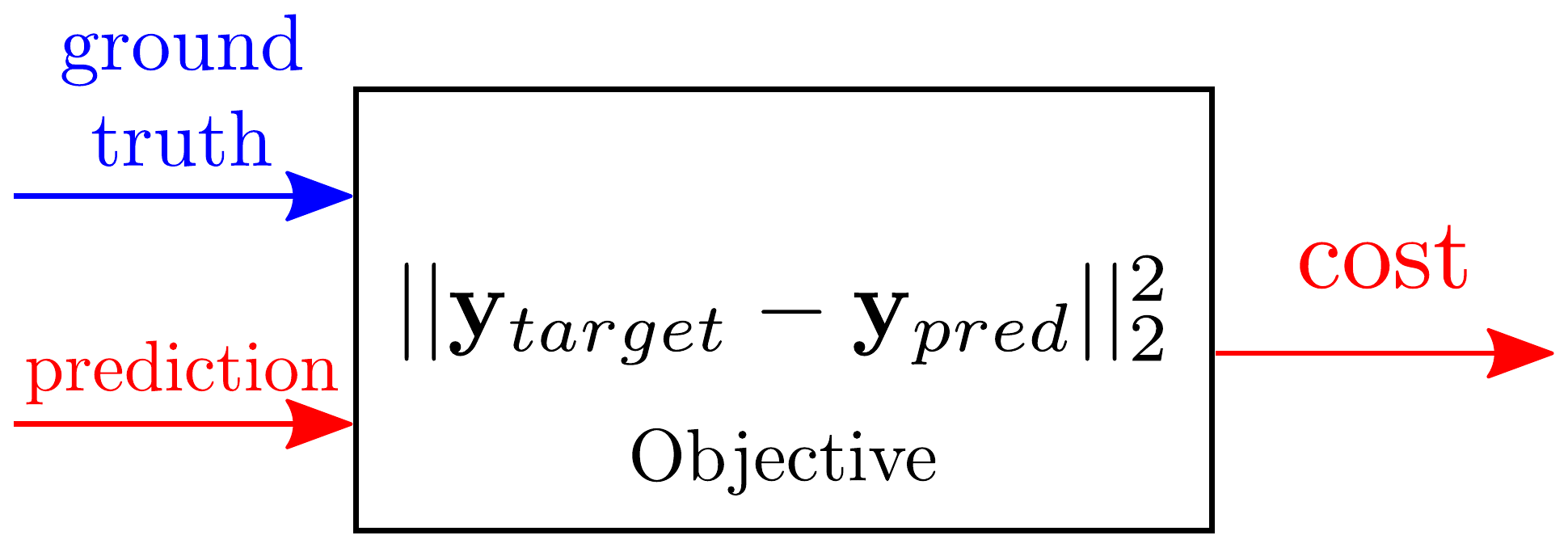}
    \caption{The symbol for the objective module. It represents the computation of a scalar cost term given prediction and ground truth inputs.}
    \label{fig:obj_mod}
\end{figure}

In all our objective modules we use the \ac{MSE} for training:  
\begin{equation}
    \mathcal{L}_{MSE} = \frac{1}{n}\sum\limits_{i=1}^n||\mathbf{y}_{target} - \mathbf{y}_{pred}||_2^2,
    \label{eq:mse}
\end{equation}
where $n$ is the number of training samples, $\mathbf{y}_{pred}$ is the model's prediction, $\mathbf{y}_{target}$ is the target value and $||.||_2^2$ is the squared Euclidean norm.
We use the \ac{MSE} even for classification, as this makes the combination of multiple objectives simpler and worked well.
Figure~\ref{fig:obj_mod} shows the symbol used in the following sections to represent an objective module.

\section{Building a Recurrent Attentive Tracking Model}
\label{sec:ratm}
The task of tracking involves mapping a sequence of input images $\mathbf{x}_1, \dots, \mathbf{x}_T$ to a sequence of object locations $\mathbf{y}_1, \dots, \mathbf{y}_T$. 
For the prediction $\hat{\mathbf{y}}_t$ of an object's location at time $t$, the trajectory $(\mathbf{y}_1, \dots, \mathbf{y}_{t-1})$ usually contains highly relevant contextual information, so it is important to choose a hidden state model which has the capacity to represent this trajectory.

\subsection{Architecture}
At each time step, the recurrent attention module outputs a glimpse from the current input frame using the attention parameters predicted at the previous time step. 
Optionally, a feature-extraction module extracts a representation of the glimpse and feeds it back to the attention module, which updates its hidden state. The tracking behavior can be learned in various ways:
\begin{itemize}
    \item One can penalize the difference between the glimpse content and a ground truth image. This can be done in the raw pixel space for simple data sets, which do not show much variation in the objects appearance. This loss is defined as
        \begin{equation}
            \mathcal{L}_{pixel} = ||\hat{\mathbf{p}} - \mathbf{p}||_2^2,
            \label{eq:pixel_loss}
        \end{equation}
        where $\hat{\mathbf{p}}$ is the glimpse extracted by the attention mechanism and $\mathbf{p}$ is the ground truth image.
        For objects with more variance in appearance, a distance measure between features extracted from the glimpse and from the ground truth image, is more appropriate:
        \begin{equation}
            \mathcal{L}_{feat} = ||f(\hat{\mathbf{p}}) - f(\mathbf{p})||_2^2,
            \label{eq:feat_loss}
        \end{equation}
        where $f(.)$ is the function computed by a feature-extraction module.

    \item Alternatively, a penalty term can also be defined directly on the attention parameters. For instance, the distance between the center of the ground truth bounding box and the
        attention mechanism's $\hat{\mathbf{g}} = (g_x, g_y)$ parameters can be used as a localization loss
        \begin{equation}
            \mathcal{L}_{loc} = ||\hat{\mathbf{g}} - \mathbf{g}||_2^2,
            \label{eq:loc_loss}
        \end{equation}
\end{itemize}
We explore several variations of this architecture in Section~\ref{sec:exp}.

\subsection{Evaluation of Tracking Performance}
\label{sec:eval}
Tracking models can be evaluated quantitatively on test data using the average \ac{IoU} \cite{everingham2010pascal}
\begin{equation}
IoU = \frac{|B_{gt}\cap B_{pred}|}{|B_{gt}\cup B_{pred}|}, 
\end{equation}
where $B_{gt}$ and $B_{pred}$ are the ground truth and predicted bounding boxes. A predicted bounding box for \ac{RATM} is defined as the rectangle between the corner points of the attention grid. This definition of predicted bounding boxes ignores the fact that each point in the glimpse is a weighted sum of pixels around the grid points and the boxes are smaller than the region seen by the attention module. While this might affect the performance under the average \ac{IoU} metric, the average \ac{IoU} still serves as a reasonable metric for the soft attention mechanism's performance in tracking.

\section{Experimental Results}
\label{sec:exp}
For an initial study, we use generated data, as described in Sections~\ref{sec:bb} and~\ref{sec:mnist}, to explore some design choices without being limited by the number of available training sequences.
In Section~\ref{sec:kth}, we show how one can apply the \ac{RATM} in a real-world context.

\subsection{Bouncing Balls}
\label{sec:bb}
For our initial experiment, we generated videos of a bouncing ball using the script released with \cite{sutskever2009recurrent}. 
The videos have $32$ frames of resolution $20\times20$.
We used $100,000$ videos for training and $10,000$ for testing.
\begin{figure}[h]
    \centering
    \includegraphics[width=.95\columnwidth]{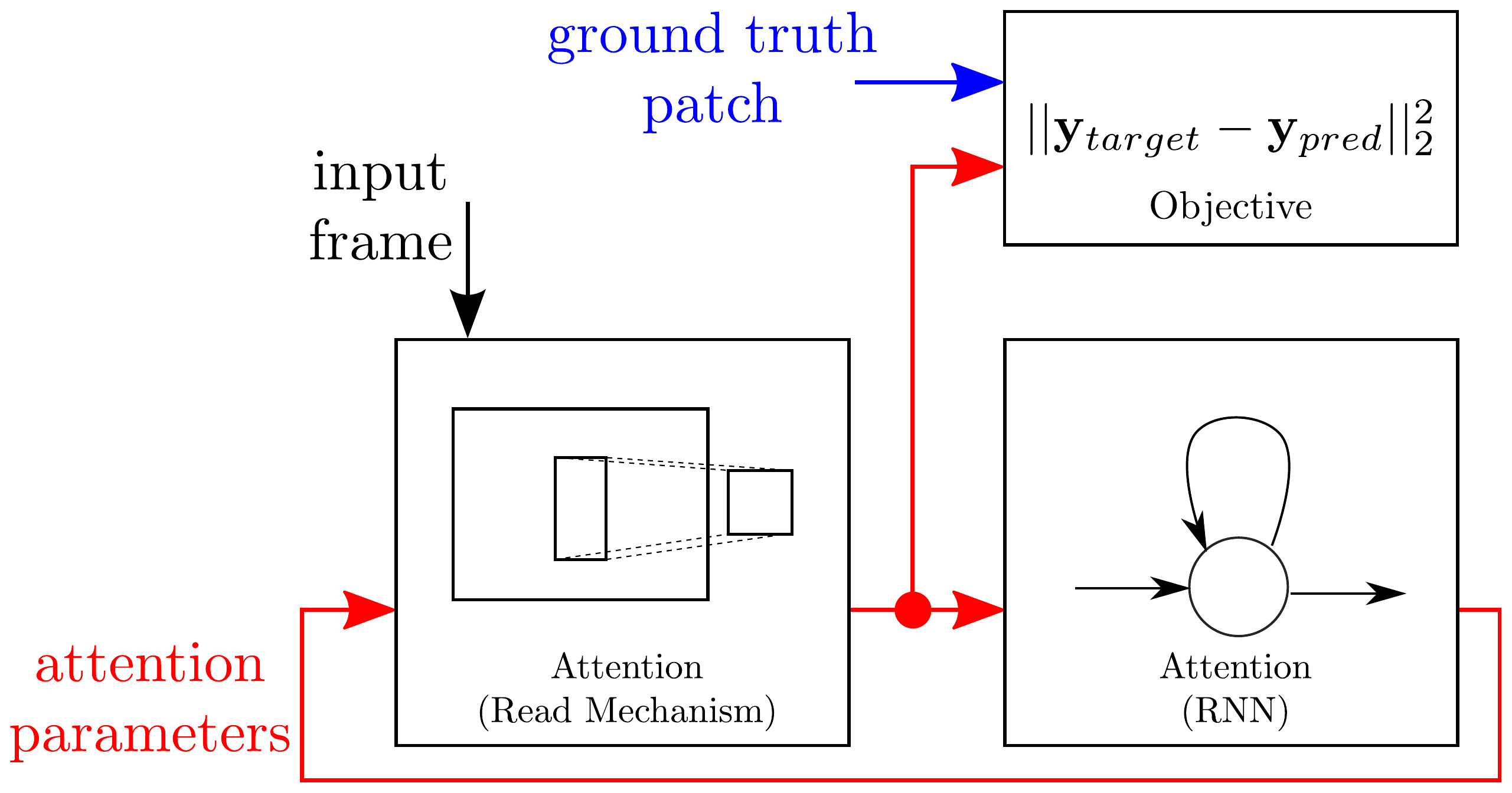}
    \caption{The architecture used for bouncing balls experiments.}
    \label{fig:bb_ratm}
\end{figure}
The \textbf{attention module} has $64$ hidden units in its \ac{RNN} and its read mechanism extracts glimpses of size $5\times5$.
The attention parameters are initialized to a random glimpse in the first frame.
The input to the \ac{RNN} are the raw pixels of the glimpse, i.e. the \textbf{feature-extraction module} here corresponds to the identity transformation.
The \textbf{objective module} computes the \ac{MSE} between the glimpse at the last time step and a target patch, which is simply a cropped white ball image, since shape and color of the object are constant across the whole data set.
Figure~\ref{fig:bb_ratm} shows a schematic of this architecture.

\begin{figure*}
    \centering
    \includegraphics[width=0.8\textwidth]{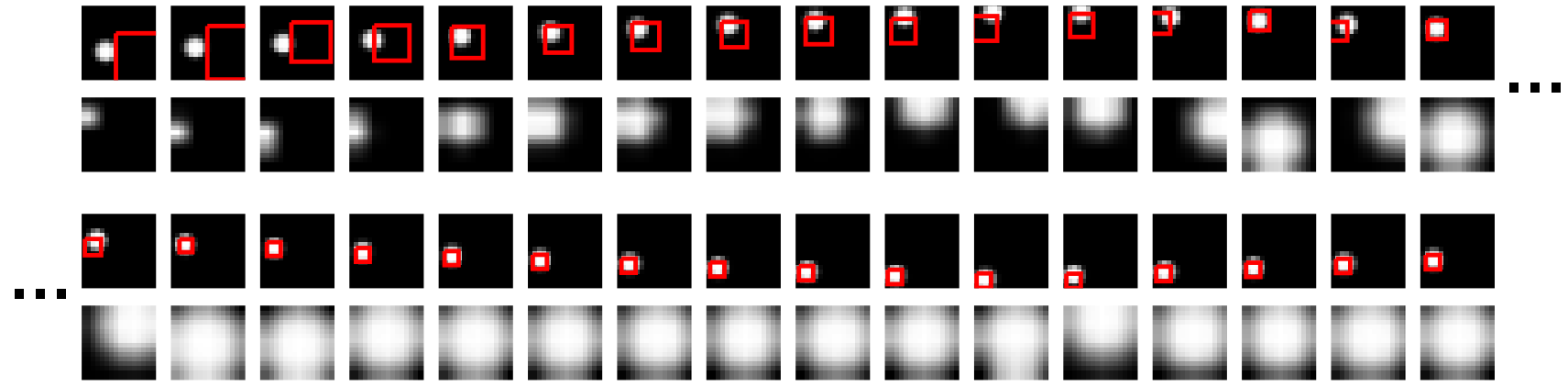}
    \caption{An example of tracking on the bouncing ball data set. The first row shows the first 16 frames of the sequence with a red rectangle indicating the location of the glimpse. The second row contains the extracted glimpses. The third and fourth row show the continuation of the sequence.}
    \label{fig:bb_seq}
\end{figure*}

\begin{figure*}
    \centering
    \includegraphics[width=0.8\textwidth]{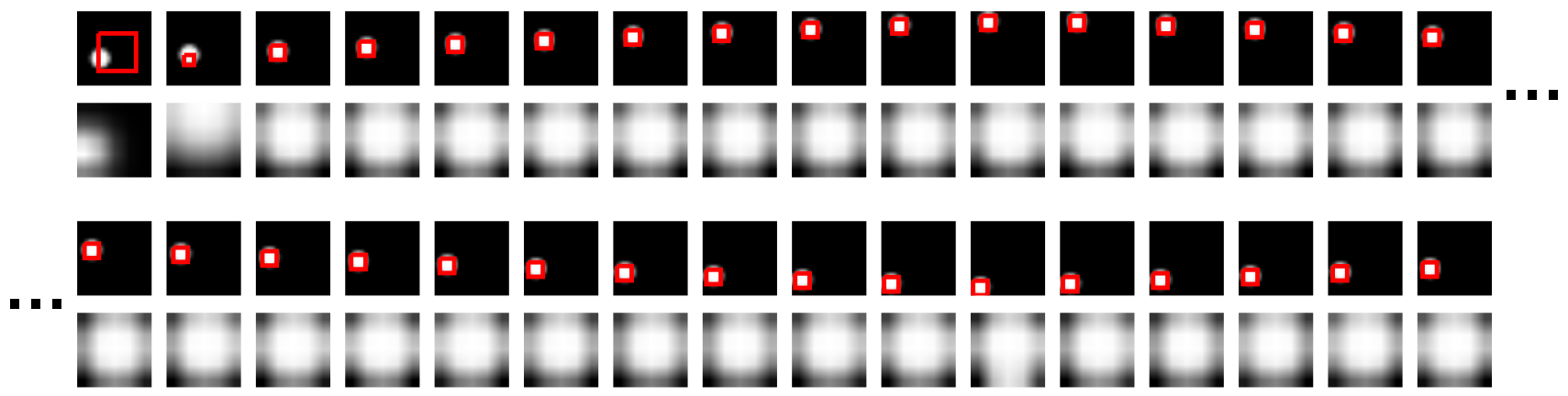}
    \caption{Tracking result on a test sequence from a model trained with the \ac{MSE} penalty at every time step. 
    The first row shows the first 16 frames of the sequence with a red rectangle indicating the location of the glimpse. 
    The second row contains the extracted glimpses. 
    The third and fourth row show the continuation of the sequence.}
    \label{fig:bb_allstep_seq}
\end{figure*}

For \textbf{learning}, we use \ac{SGD} with a mini-batch size of $16$, a learning rate of $0.01$ and gradient clipping \cite{pascanu2012difficulty} with a threshold of $1$ for $200$ epochs.
Figure~\ref{fig:bb_seq} shows results of tracking a ball in a test sequence.
\ac{RATM} is able to learn the correct tracking behaviour only using the penalty on the last frame.
We also trained a version with the objective module computing the average \ac{MSE} between glimpses of all time steps and the target patch. 
An example tracking sequence of this experiment is shown in Figure~\ref{fig:bb_allstep_seq}.
The first two rows of Table~\ref{tab:iou} show the test performance of the model trained with only penalizing the last frame during training. 
The first row shows the average \ac{IoU} of the last frame and the second shows the average \ac{IoU} over all 32 frames of test sequences.
The third row shows the average \ac{IoU} over all frames of the model trained with the penalty on all frames.

The model trained with the penalty at every time step is able to track a bouncing ball for sequences that are much longer than the training sequences. 
We generated videos that are almost ten times longer (300 frames) and \ac{RATM} reliably tracks the ball until the last frame.  
An example is uploaded as part of the supplementary material.

The dynamics in this data-set are rather limited, but as a proof-of-concept they show that the model is able to learn tracking behavior end-to-end.
We describe more challenging tasks in the following sections.

\subsection{MNIST}
\label{sec:mnist}
To increase the difficulty of the tracking task, we move to more challenging data sets, containing more than a single type of object (ten digits), each with variation. We generate videos from $28\times28$ MNIST images of handwritten digits \cite{lecun1998gradient} by placing randomly-drawn digits in a larger $100\times100$ canvas with black background and moving the digits from one frame to the next.
We respected the same data split for training and testing as in the original MNIST data-set, i.e. digits were drawn from the training split to generate training sequences and from the test split for generation of the test sequences. 

\begin{figure}[h]
    \centering
    \includegraphics[width=.95\columnwidth]{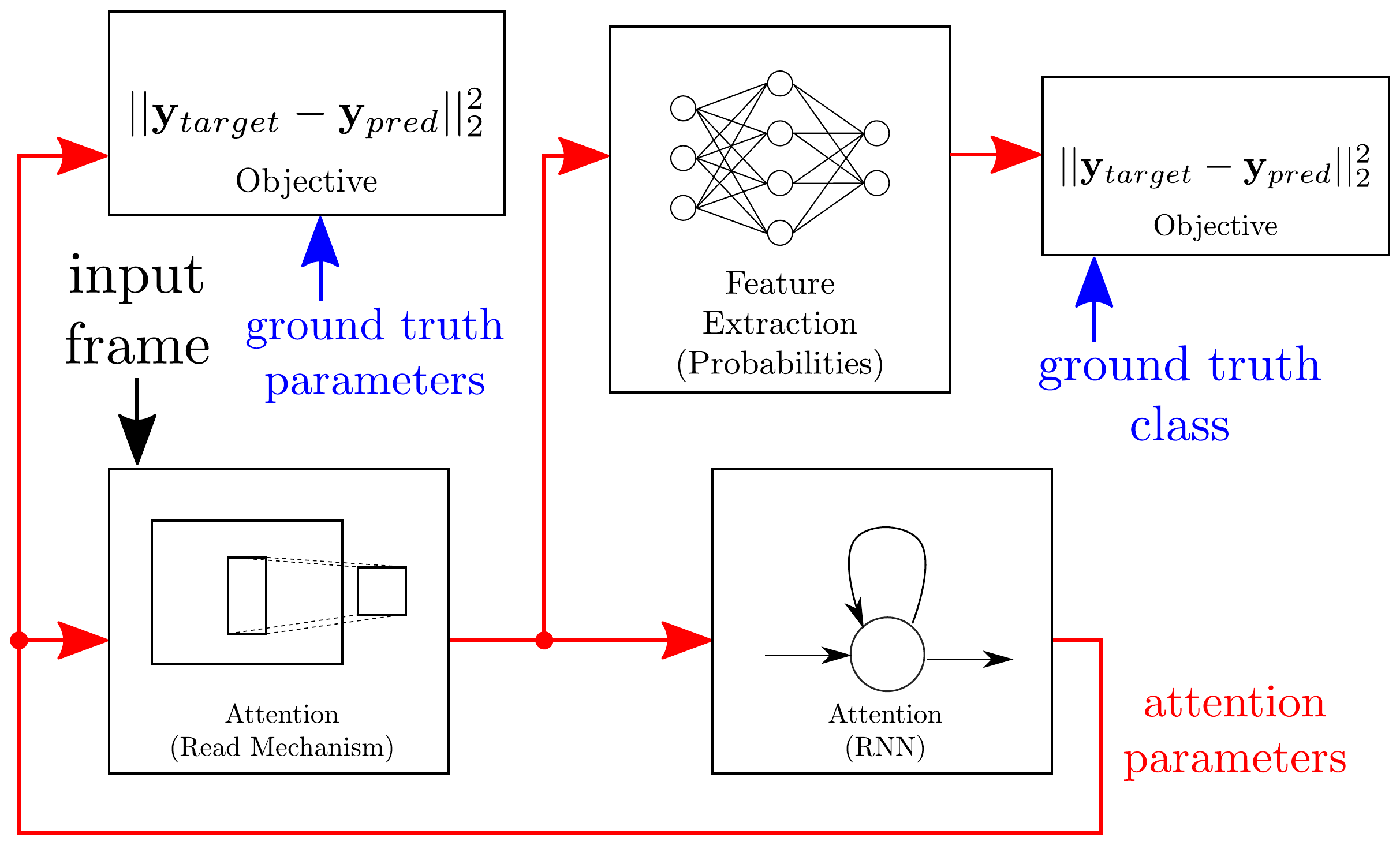}
    \caption{The architecture used for MNIST experiments.}
    \label{fig:mnist_ratm}
\end{figure}

Figure~\ref{fig:mnist_ratm} shows the schematic of \ac{RATM} for the MNIST experiments.
The \textbf{attention module} is similar to the one used in Section~\ref{sec:bb}, except that its \ac{RNN} has $100$ hidden units and the size of the glimpse is $28\times28$ (the size of the MNIST images and the \ac{CNN} input layer).

In the bouncing balls experiment we were able to generate a reliable training signal using pixel-based similarity.
However, the variation in the MNIST data set requires a representation that is robust against small variations to guide the training. 
For this reason, our \textbf{feature-extraction module} consists of a (relatively shallow) \ac{CNN}, that is pre-trained on classification of MNIST digits.
Note, that this \ac{CNN} is only used during training.
The \ac{CNN} structure has two convolutional layers with filter bank sizes of $32\times5\times5$, each followed by a $2\times2$ maxpooling layer, $0.25$ dropout \cite{hinton2012improving}, and the \ac{ReLU} activation function. These layers are followed by a 10-unit softmax layer for classification. 
The \ac{CNN} was trained using \ac{SGD} with a mini-batch size of $128$, a learning rate $0.01$, momentum of $0.9$ and gradient clipping with a threshold of $5.0$ to reach a validation accuracy of $99\%$.

This \ac{CNN} classifier is used to extract class probabilities for each glimpse and its parameters remain fixed after pre-training.
One of the \textbf{objective modules} computes the loss using these probabilities and the target class.
Since training did not converge to a useful solution using only this loss, we first introduced an additional objective module penalizing the distances between the upper-left and lower-right bounding-box corners and the corresponding target coordinates. 
While this also led to unsatisfactory results, we found that replacing the bounding box objective module with one that penalized only grid center coordinates worked well. 
One possible explanation is, that the grid center penalty does not constrain the stride. 
Therefore, the glimpse is free to explore without being forced to zoom in.
The two penalties on misclassification and on grid center distance, helped the model to reliably find and track the digit. 
The localization term helped in the early stages of training to guide \ac{RATM} to track the digits, whereas the classification term encourages the model to properly zoom into the image to maximize classification accuracy.

For \textbf{learning} we use \ac{SGD} with mini-batch size of $32$, a learning rate of $0.001$, momentum of $0.9$ and gradient clipping with a threshold of $1$ for $32,000$ gradient descent steps.

\subsubsection{Single-Digit}
In the first MNIST experiment, we generate videos, each with a single digit moving in a random walk with momentum.
The data set consists of $100,000$ training sequences and $10,000$ test sequences.
The initial glimpse roughly covers the whole frame.

\begin{figure*}
    \centering
    \includegraphics[width=0.8\textwidth]{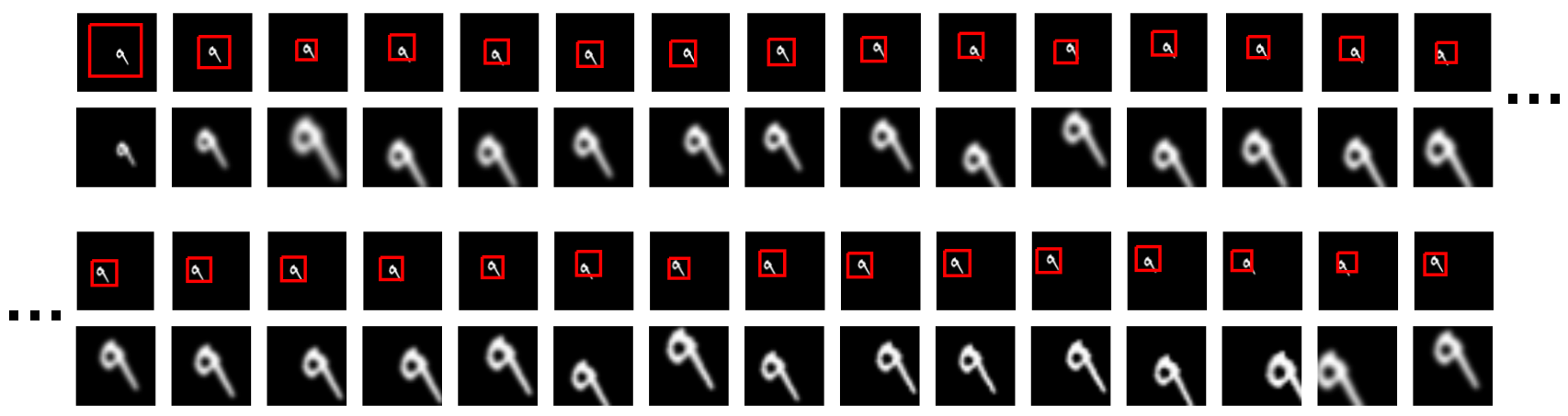}
    \caption{Tracking one digit. The first and second row show the sequence and corresponding extracted glimpses, respectively. 
    The red rectangle indicates the location of the glimpse in the frame.
    The third and fourth row are the continuation. Prediction works well far beyond the training horizon of 10 frames.}
    \label{fig:singlemnist_seq}
\end{figure*}

Training is done on sequences with only $10$ frames. The classification and localization penalties were applied at every time-step.
At test time, the \ac{CNN} is switched off and we let the model track test sequences of $30$ frames.
The fourth row of Table~\ref{tab:iou} shows the average \ac{IoU} over all frames of the test sequences.
Figure~\ref{fig:singlemnist_seq} shows one test sample.

\subsubsection{Multi-Digit}
\label{sec:multidigit}
It it interesting to investigate how robust \ac{RATM} is in presence of another moving digit in the background.
To this end, we generated new sequences by modifying the bouncing balls script released with \cite{sutskever2009recurrent}. 
The balls were replaced by randomly drawn MNIST digits.
We also added a random walk with momentum to the motion vectors.
We generated $100,000$ sequences for training and $5,000$ for testing.

\begin{figure*}
    \centering
    \includegraphics[width=0.8\textwidth]{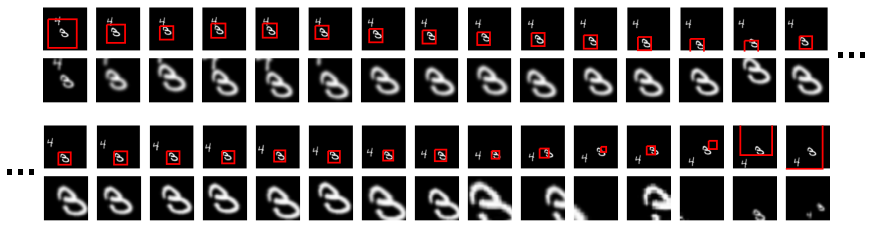}
    \caption{Tracking one of two digits. The first and second row show the sequence and corresponding extracted glimpses, respectively. The red rectangle indicates the location of the glimpse in the frame.
    The third and fourth row are the continuation. Prediction works well for sequences twice as long as the training sequences with 10 frames.}
    \label{fig:multimnist_seq}
\end{figure*}

Here, the bias for attention parameters is not a learn-able parameter. 
For each video, the bias is set such that the initial glimpse is centered on the digit to be tracked. 
Width and height are set to about $80\%$ of the frame size.
The model was also trained on $10$ frame sequences and was able to focus on digits for at least $15$ frames on test data.
Figure~\ref{fig:multimnist_seq} shows tracking results on a test sequence.
The fifth row of Table~\ref{tab:iou} shows the average \ac{IoU} of all test sequences over $30$ frames.

\subsection{Tracking humans in video}
\label{sec:kth}
To evaluate the performance on a real-world data set, we train \ac{RATM} to track humans in the KTH action recognition data set \cite{KTH}, which has a reasonably large number of sequences.
We selected the three activity categories, which show considerable motion: walking, running and jogging.
We used the bounding boxes provided by \cite{KTHBoundingBoxes}, which were not hand-labeled and contain noise, such as bounding boxes around the shadow instead of the subject itself.

For the \textbf{feature-extraction module} in this experiment, we trained a \ac{CNN} on binary~--~human vs. background~--~classification of $28\times28$ grayscale patches.
To generate training data for this \ac{CNN}, we cropped positive patches from annotated subjects in the ETH pedestrian \cite{ETHPedestrians} and INRIA person \cite{INRIAPerson} data sets. 
Negative patches were sampled from the KITTI detection benchmark \cite{Geiger2012CVPR}. 
This yielded $21,134$ positive and $29,923$ negative patches, of which we used $20,000$ per class for training.
The architecture of the \ac{CNN} is as follows: two convolutional layers with filter bank sizes $128\times5\times5$ and $64\times3\times3$, each followed by $2\times2$ max-pooling and a \ac{ReLU} activation. 
After the convolutional layers, we added one fully-connected \ac{ReLU}-layer with $256$ hiddens and the output softmax-layer of size $2$.
For \textbf{pre-training}, we used \ac{SGD} with a mini-batch size of $64$, a learning rate of $0.01$, momentum of $0.9$ and gradient clipping with a threshold of $1$. 
We performed early stopping with a held-out validation set sampled randomly from the combined data set.

As this real-world data set has more variation than the previous data sets, the \textbf{attention module}'s \ac{RNN} can also benefit from a richer feature representation. 
Therefore, the \ac{ReLU} activations of the second convolutional layer of the feature-extraction module are used as input to the attention module.
The \ac{RNN} has $32$ hidden units. 
This low number of hidden units was selected to avoid overfitting, as the number of sequences ($1,200$ short sequences) in this data set is much lower than in the synthetic data sets. 
We initialize the attention parameters for the first time step with the first frame's target window.
The initial and target bounding boxes are scaled up by a factor of $1.5$ and the predicted bounding boxes are scaled back down with factor $\frac{1}{1.5}$ for testing. 
This was necessary, because the training data for the feature-extraction module had significantly larger bounding box annotations.

\begin{figure*}
    \centering
    \includegraphics[width=.7\textwidth]{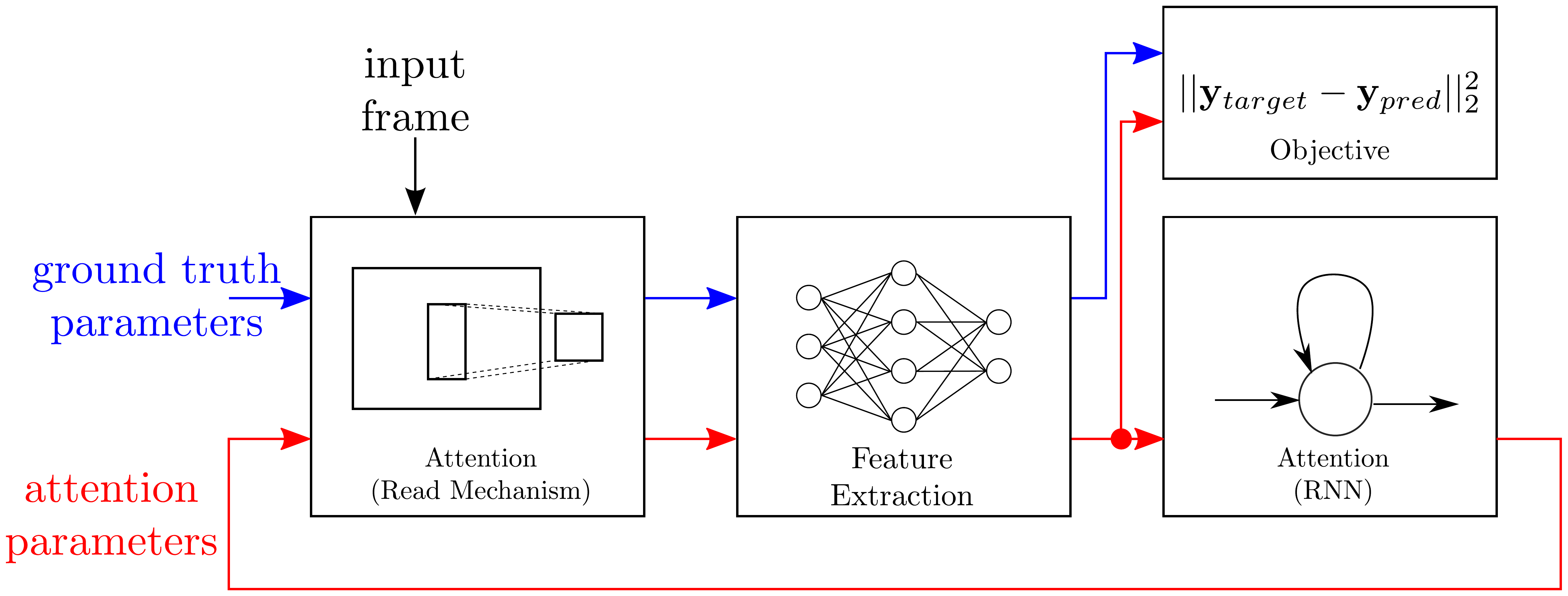}
    \caption{The architecture used for KTH experiments.}
    \label{fig:kth_ratm}
\end{figure*}

The inputs to the \textbf{objective module} are the \ac{ReLU} activations of the fully-connected layer, extracted from the predicted window and from the target window.
The computed cost is the \ac{MSE} between the two feature vectors.
We also tried using the cosine distance between two feature vectors, but did not observe any improvement in performance.
The target window is extracted using the same read mechanism as in the attention module. 
Simply cropping the target bounding boxes would have yielded local image statistics that are too different from windows extracted using the read mechanism.
Figure~\ref{fig:kth_ratm} shows the schematic of the architecture used in this experiment.

For \textbf{learning}, we used \ac{SGD} with a mini-batch size of $16$, a learning rate of $0.001$ and gradient clipping with a threshold of $1.0$. 
In this experiment we also added a weight-decay regularization term to the cost function that penalizes the sum of the squared Frobenius norms of the \ac{RNN} weight matrices from the input to the hidden layer and from the hidden layer to the attention parameters. 
The squared Frobenius norm is defined as 
\begin{equation}
    ||\mathbf{A}||_F^2 = \sum\limits_i^m\sum\limits_j^n |a_{ij}|^2,
\end{equation}
where $a_{ij}$ is the element at row $i$, column $j$ in matrix $\mathbf{A}$.
This regularization term improved the stability during learning.
As another stabilization measure, we started training with short five-frame sequences and increased the length of sequences by one frame every $160$ gradient descent steps.

For evaluation, we performed a leave-one-subject-out experiment. 
For each of the $25$ subjects in KTH, we used the remaining $24$ for training and validation. 
A validation subject was selected randomly and used for early stopping.
The reported number in the sixth row of Table~\ref{tab:iou} is the \ac{IoU} on full-length videos of the test subject averaged over all frames of each left-out subject and then averaged over all subjects.

% test-person 25, sample from class jogging
\begin{figure*}%[h!]
    \centering
    \includegraphics[width=0.9\textwidth]{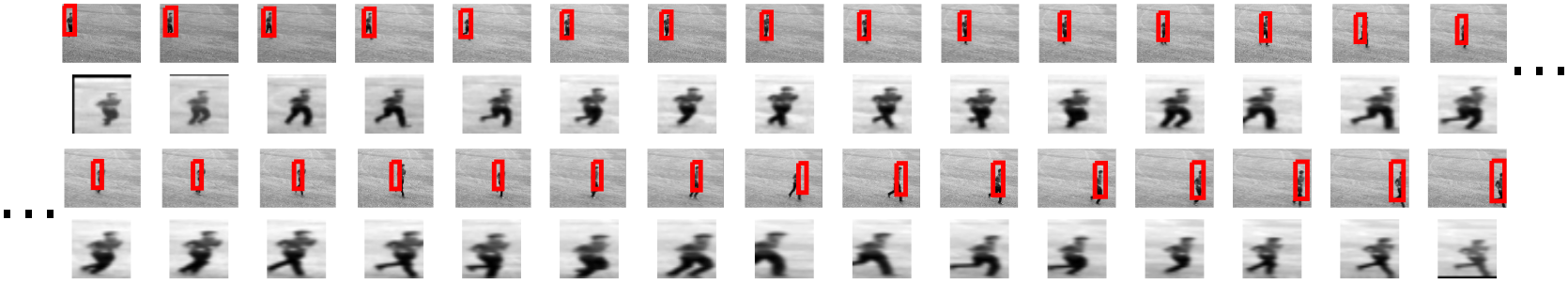}
    
    \vspace{-3pt}
    \rule{.8\textwidth}{1pt}
    \vspace{7pt}
    
% test-person 1, sample from class walking
    \includegraphics[width=0.9\textwidth]{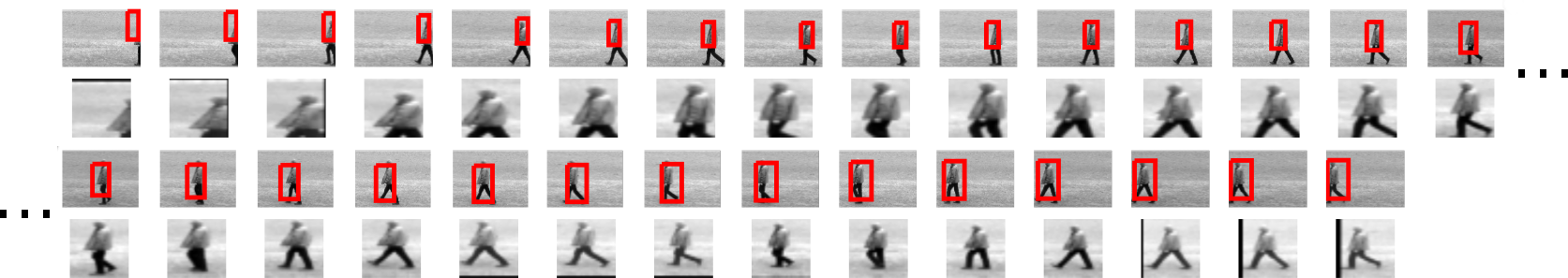}
    \caption{Two examples of tracking on the KTH data set. The layout for each example is as follows: the first row shows 15 frames of one test sequence with a red rectangle indicating the location of the glimpse. The second row contains the extracted glimpses. 
    The third and fourth row show the continuation of the sequence.
    We only show every second frame.}
    \label{fig:kth_seq}
\end{figure*}
Figure~\ref{fig:kth_seq} shows examples of test sequences for the classes \emph{jogging} and \emph{walking}.
Note, that the region captured by the glimpses is larger than the bounding boxes, because the model internally scales the width and height by factor $1.5$ and the Gaussian sampling kernels of the attention mechanism extend beyond the bounding box.
An interesting observation is that \ac{RATM} scales up the noisy initial bounding box in Figure~\ref{fig:kth_seq} (bottom example), which covers only a small part of the subject.
This likely results from pre-training the feature-extraction module on full images of persons. 
We observed a similar behavior for multiple other samples.
Although the evaluation assumes that the target bounding boxes provided by \cite{KTHBoundingBoxes} are accurate, \ac{RATM} is able to recover from such noise.

\begin{figure*}
    \centering
    \includegraphics[width=.99\textwidth]{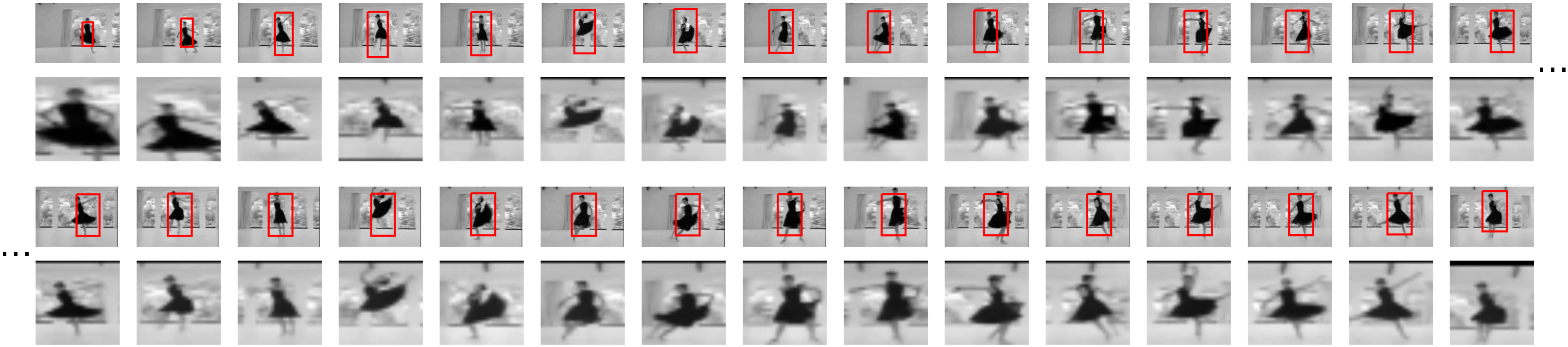} %sampled every 7 frames
    
    \vspace{-3pt}
    \rule{.8\textwidth}{1pt}
    \vspace{7pt}
    
    \includegraphics[width=.99\textwidth]{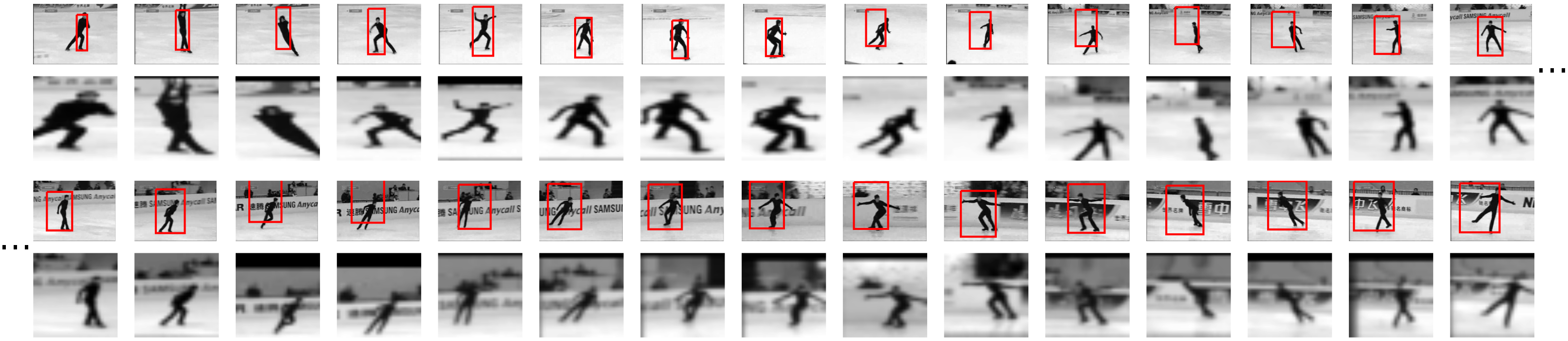} %sampled every 10 frames
    
    \vspace{-3pt}
    \rule{.8\textwidth}{1pt}
    \vspace{7pt}
    
    \includegraphics[width=.99\textwidth]{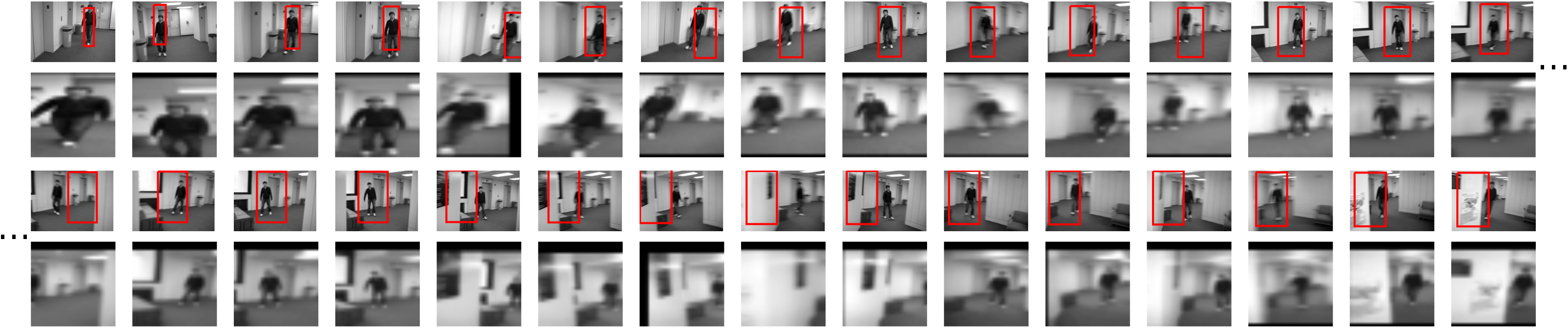} %sampled every 10 frames
    
    \vspace{-3pt}
    \rule{.8\textwidth}{1pt}
    \vspace{7pt}
    
    \includegraphics[width=.99\textwidth]{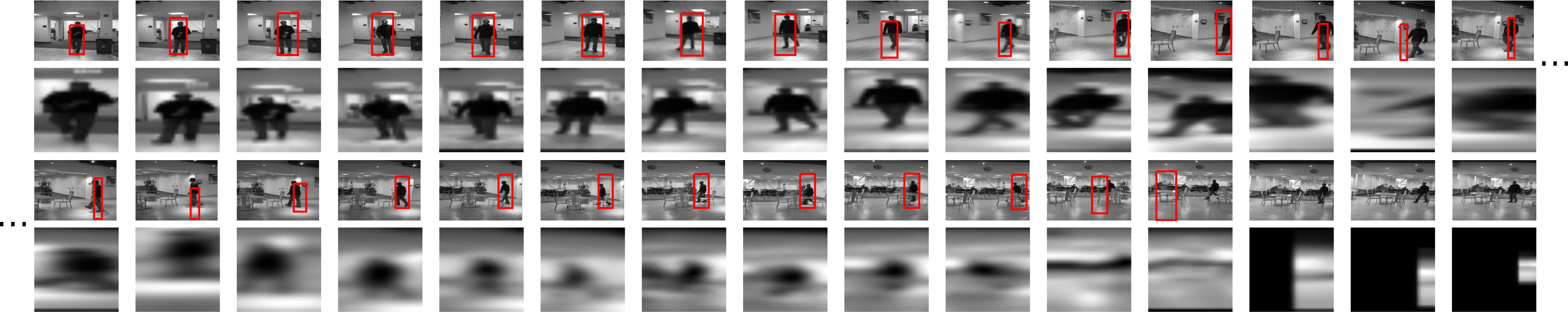} %sampled every 10 frames
    \caption{Predictions of a KTH model on sequences from the TB-100 benchmark. From top to bottom we show the sequences \emph{Dancer}, \emph{Skater2}, \emph{BlurBody} and \emph{Human2}. To save space, we only show every seventh frame of the Dancer predictions and every tenth frame of the other sequences. 
    The layout for each sequence is as follows:
    The first row shows 15 frames of one test sequence with a red rectangle indicating the location of the predicted glimpse. The second row contains the extracted glimpses. 
    The third and fourth row show the continuation of the sequence.}
    \label{fig:kth_model_on_tb100}
\end{figure*}

To show how the model generalizes to unseen videos containing humans, we let it predict some sequences of the TB-100 tracking benchmark \cite{wu2015object}.
For this experiment, we picked one of the $25$ KTH model, that had a reasonably stable learning curve (\ac{IoU} over epochs). % no 20
As an example, Figure~\ref{fig:kth_model_on_tb100} shows every seventh predicted frame of the \emph{Dancer} sequence and every tenth predicted frame of the sequences \emph{Skater2}, \emph{BlurBody} and \emph{Human2}.
For the first two examples, \emph{Dancer} and \emph{Skater2}, \ac{RATM} tracks the subjects reliably through the whole length of the sequence.
This is interesting, as the tracking model was only trained on sequences of up to $30$ frames length and the variation in this data is quite different from KTH.
The BlurBody and Human2 sequences are more challenging, including extreme camera motion and/or occlusions, causing the model to fail on parts of the sequence.
Interestingly in some cases it seems to recover. 

In general, the model shows the tendency to grow the window, when it loses a subject. 
This might be explained by instability of the \ac{RNN} dynamics and blurry glimpses due to flat Gaussians in the attention mechanism. 
These challenges will be discussed further in Section~\ref{sec:ratm_discussion}.

\begin{table*}
\caption{Average Intersection-over-Union scores on test data.}
\label{tab:iou}
\begin{center}
\begin{tabular}{ll}
\multicolumn{1}{c}{\bf Experiment}  &\multicolumn{1}{c}{\bf Average \ac{IoU} (over \# frames)}
\\ \hline \\
Bouncing Balls (training penalty only on last frame)             &69.15 (1, only last frame)\\
Bouncing Balls (training penalty only on last frame)             &54.65 (32)\\ 
Bouncing Balls (training penalty on all frames)            &66.86 (32)\\
MNIST (single-digit)             &63.53 (30)\\
MNIST (multi-digit)             &51.62 (30)\\
KTH (average leave-one-subject-out) & 55.03 (full length of test sequences)
\end{tabular}
\end{center}
\end{table*}

\section{Discussion}
\label{sec:ratm_discussion}
We propose a novel neural framework including a soft attention mechanism for vision, and demonstrate its application to several tracking tasks. 
Contrary to most existing similar approaches, \ac{RATM} only processes a small window of each frame.
The selection of this window is controlled by a learned attentional behavior. 
Our experiments explore several design decisions that help overcome challenges associated with adapting the model to new data sets.
Several observation in the real-world scenario in Section~\ref{sec:kth}, are important for applications of attention mechanisms in computer vision in general: 
\begin{itemize}
    \item The model can be trained on noisy bounding box annotation of videos and at test time recover from noisy initialization.
        This might be related to the pre-training of the feature-extraction module on static images.
        The information about the appearance of humans is transferred to the attention module, which learns to adapt the horizontal and vertical strides among other parameters of the glimpse to match this appearance.
    \item The trained human tracker seems to generalize to related but more challenging data.
\end{itemize}

\section{Directions for Future Research}
The modular neural architecture is fully differentiable, allowing end-to-end training.
End-to-end training allows the discovery of spatio-temporal patterns, which would be hard to learn with separate training of feature extraction and attention modules.
In future work we plan to selectively combine multiple data sets from different tasks, e.g. activity recognition, tracking and detection.
This makes it possible to benefit from synergies between tasks \cite{caruana1997multitask}, and can help overcome data set limitations. 

One could also try to find alternatives for the chosen modules, e.g. replacing the read mechanism with \emph{spatial transformers} \cite{jaderberg2015spatial}.
Spatial transformers offer a more general read mechanism, that can learn to align glimpses using various types of transformations.
The application of Spatial Transformers in \acp{RNN} for digit recognition has been explored in \cite{sonderby2015recurrent}.

\section*{Acknowledgments}
The authors would like to thank the developers of Theano \cite{bastien2012theano,bergstra2010theano}. We thank Kishore Konda, Jörg Bornschein and Pierre-Luc St-Charles for helpful discussions.
This work was supported by an NSERC Discovery Award, CIFAR, FQRNT and the German BMBF, project 01GQ0841.

\ifCLASSOPTIONcaptionsoff
  \newpage
\fi

\bibliographystyle{IEEEtran}
\bibliography{IEEEabrv,literature}

\end{document}